\documentclass[runningheads]{llncs}
\usepackage{graphicx}
\usepackage{tikz}
\usepackage{blindtext}
\usepackage{comment}
\usepackage{amsmath,amssymb} 
\usepackage{color}
\usepackage{float}

\usepackage[accsupp]{axessibility}  
\usepackage{cite}
\usepackage{marvosym}

\begin{document}

\pagestyle{headings}
\mainmatter

\title{Content-Oriented Learned Image Compression} 

\titlerunning{Content-Oriented Learned Image Compression}

\author{Meng Li \and
Shangyin Gao \and
Yihui Feng \and
Yibo Shi \and
Jing Wang \textsuperscript{\Letter}}

\authorrunning{M. Li et al.}

\institute{Huawei Technologies, Beijing, China \\
\email{wangjing215@huawei.com}}

\maketitle

\begin{abstract}
In recent years, with the development of deep neural networks, end-to-end optimized image compression has made significant progress and exceeded the classic methods in terms of rate-distortion performance. However, most learning-based image compression methods are unlabeled and do not consider image semantics or content when optimizing the model. In fact, human eyes have different sensitivities to different content, so the image content also needs to be considered. In this paper, we propose a content-oriented image compression method, which handles different kinds of image contents with different strategies.  Extensive experiments show that the proposed method achieves  competitive subjective results compared with state-of-the-art end-to-end learned image compression methods or classic methods. 

\end{abstract}

\section{Introduction}
\begin{figure}[t]
    \centering
    \includegraphics[width=0.9\textwidth]{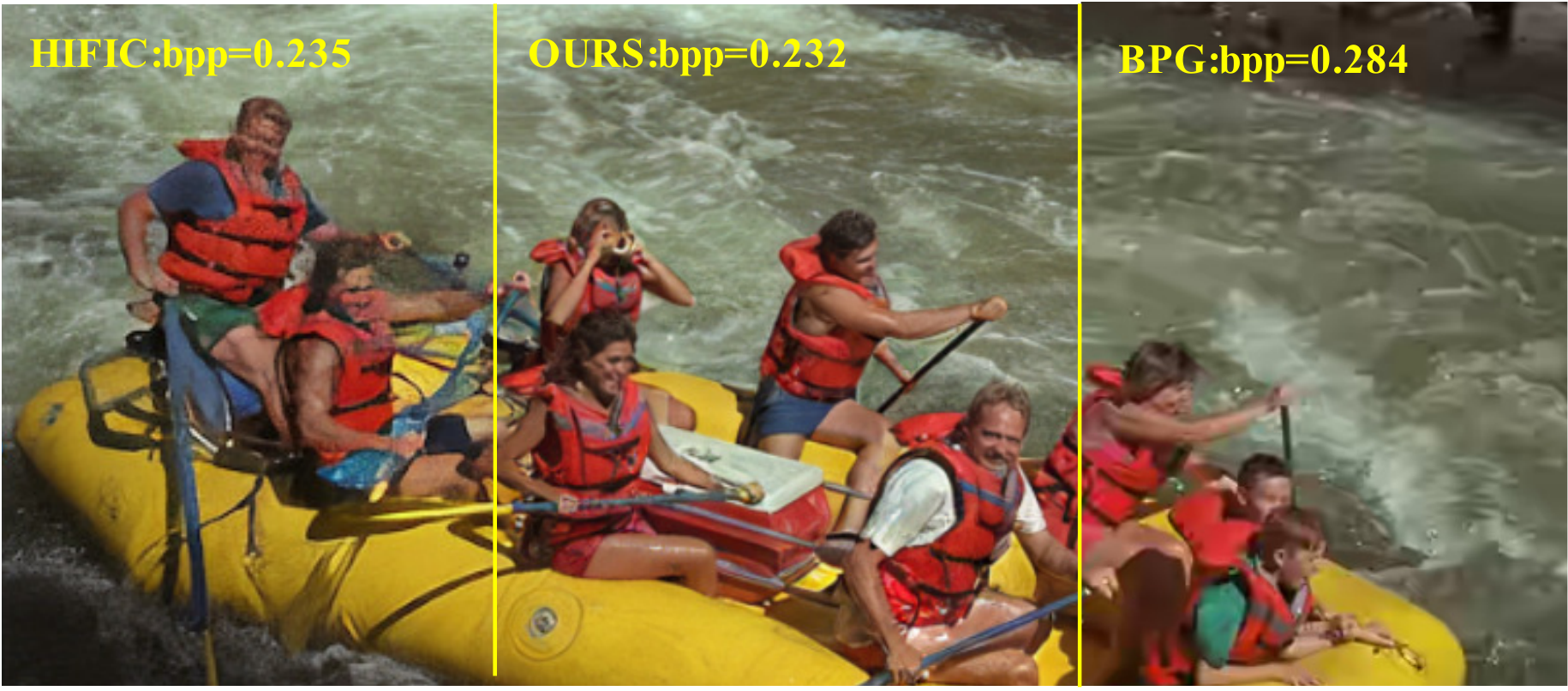}
    \caption{Reconstructed image with different methods at a similar bitrate.}
    \label{fig:intro}
\end{figure}

An uncompressed image usually contains tremendous data that is expensive to store or transmit. Therefore, image compression, or image codec, which aims to reduce redundant information in image data for efficient storage and transmission, is an essential part of real-world applications. The frameworks of traditional image compression methods such as JPEG \cite{JPEG}, JPEG2000 \cite{JPEG2000} and BPG \cite{BPG} are sophistically designed, which usually include modules of prediction, transformation, quantization, and entropy coding. However, each module in the frameworks is optimized separately, making it hard to achieve global optimality.  \\
\indent
In recent years, with the development of deep learning, end-to-end (E2E) learned image compression methods are proposed. Compared with traditional codecs, the biggest advantage of E2E methods is that the whole framework can be jointly optimized, making it has a greater protential in compression efficiency. Impressively, it takes only five years for E2E image compression to outpace the traditional methods that have developed for 30 years in terms of rate-distortion performance. Mainstream E2E image compression methods rely on the variational autoencoder(VAE) \cite{Bella2016}\cite{Bella2018}\cite{LeeCTX}\cite{BellaCTX} or generative adversarial network (GAN) \cite{AgustssonGAN}\cite{AkutsuGAN}\cite{HIFIC}\cite{RippelGAN} to compress images, and various loss metrics such as Mean Squared Error (MSE), Multi-Scale Structural Similarity Index (MS-SSIM) \cite{MSSSIM}, Learned Perceptual Image Patch Similarity (LPIPS) \cite{LPIPS}, etc. are used to optimize the model. Most existing methods optimized all regions of the image in the same way, and have different distortions at low bit rates. For example, Minnen \cite{BellaCTX} used MSE metric to train a VAE image compression network. In this optimization, structural information such as text and lines are correctly perserved, but the reconstructed image will become too blurry. Mentzer \cite{HIFIC} tried to improve the image quality by introducing GAN and LPIPS. In this optimization, the informative details are well preserved, but the structural information is distorted, such as distorted text, warped face (demonstrate in the left part of Fig.\ref{fig:intro}).  \\
\indent
Image content plays an important role in human perception, and human eyes have different sensitivities to different content. However, the influence of image content has been largely ignored in learning-based image compression. Different regions of the images have different properties, so the training strategy should also be different. For example, pixels in flat or texture regions are strongly correlated, and it is better to use loss metrics with a large receptive field. In contrast, pixels in edge or structure reagions have little correlation with their neighborhoods, and it is better to use loss metrics with a small receptive field. The usage of paintbrushes is a good analogy to describe it, large paintbrushes are usually used to make large textures and to lay large color blocks, while small paintbrushes are usually used to draw fine lines and color dots.  Besides, for some special regions, like small face, it is needed to use more strict constraints to avoid deformation. As a result, to address the problems above, we propose a content-oriented image compression scheme, which is suitable for most of the existing frameworks. Specifically, in the training stage, we divide the image into different regions, and use different loss metrics according to their characteristics. \\
\indent
To the best of our knowledge, this is the first content-oriented image compression method that improves the visual perceptual quality without changing the architecture. An instance  is shown in Fig. \ref{fig:intro}, which demonstrates the superiority of our method.  The contributions of this paper are summarized as follows:
\begin{enumerate}
\item We propose a content-oriented E2E image compression scheme, in which we use different loss metrics in different regions. The region masks are used only in the training stage, and no extra information is needed in the encoding and decoding stage.
\item To evaluate the scheme, we design a GAN-based architecture as an instance to show the effectiveness. Several classic and E2E methods are used for comparison, and our method shows the best performance.
\end{enumerate}

\section{Related Work}
\subsection{Loss metrics}
\subsubsection{MAE/MSE Loss} MSE and the mean absolute error (MAE) are two of the most commonly used loss metrics in image compression. MAE and MSE metrics assume pixel-wise independence, and constrints the accuracy of the corresponding pixels. However, the main drawback of MSE loss metric mostly yields an oversmoothed reconstruction, which results in a lack of high-frequency details in the edges and textures. Given an image $x$ and its reconstructed version $\tilde{x}$, the MAE and MSE loss are computed by:
\begin{equation}
\label{mse}
\begin{aligned}
\mathcal{L}_{m a e} &=\mathbb{E}\|x-\tilde{x}\|_{1} \\
\mathcal{L}_{m s e} &=\mathbb{E}\|x-\tilde{x}\|_{2}^2
\end{aligned}
\end{equation}

\subsubsection{Laplacian Loss} Laplacian is a differential operator given by the divergence of the gradient, which is always used to detect the high-frenquency component of images. The Laplacian loss is defined as the mean-squared distance between their Laplacians:
\begin{equation}
\label{lap}
\mathcal{L}_{{lap }}=\mathbb{E}\|L\left(x\right)-L\left(\tilde{x}\right)\|_{2}^2
\end{equation}
Where $\mathcal{L}_{\text {lap }}$ is the Laplacian loss value. As shown in Eq.\ref{lap}, it is also a point-to-point loss metric, which constraints the Laplacians of the corresponding location. By doing this, the high-frenquency component of the area is preserved. \\
\subsubsection{LPIPS Loss} LPIPS is a state-of-the-art perceptual metric proposed by Zhang \cite{LPIPS}. He uses architectures trained for classification as feature extractors, and leverages the power of the features to judge the similarity between the reconstructed image and the original one, which is computed by:
\begin{equation}
\label{lpips}
\mathcal{L}_{l p i p s}=\sum_{k} \tau^{k}\left(f^{k}\left(x\right)-f^{k}\left(\tilde{x}\right)\right) 
\end{equation}
where $f$ denotes the feature extractor, and $\tau^{k}$ computes the score of features from the k-th layer of the architecture. The LPIPS value is the averaged score of all layers. It is a loss metric with a large receptive field, which constraints the distribution of corresponding location. The LPIPS metrics makes images more semantically similar, which is more consistent with human perception. However, the absence of point to point constraints may result in geometric distortion.  \\
\subsubsection{GAN Loss} GAN is widely used for perceptual image compression. By taking advantage of adversarial training, GAN-based architectures can produce more photo-realistic images. It contains two rivaling networks: the generator G is used to generate an image $\tilde{x}=G(\hat{y})$ that is consistent with the input image distribution $p_{\boldsymbol{x}}$, and the discriminator D is used to predict if the input image is an image generated by G. The goal of GAN for G is to produce images that are real enough to fool D. In the procedure of training, the image produced by G becomes more and more authentic, and D becomes more and more discriminating, finally reaching a balance. 
\subsection{learned image compression methods}
Many E2E image compression methods take advantage of VAE as its backbone, which usually consists of four components: the encoder, the quantizer, the entropy module and the decoder. The encoder is used to encode the input image $x$ into a latent representation $y$, which is then quantized to $\hat{y}$ by the quantizer. The entropy module, which is used to estimate the distribution of $\hat{y}$, plays an important role in minimizing the rate-distortion cost. Finally, the quantified feature $\hat{y}$ is transformed to reconstruct the image $\tilde{x}$. The framework is directly optimized for the rate-distortion trade-off, that is: $\mathcal{L}_{R D}=R+\lambda d$. Where \textsl{R} represents the bitrate which is lower-bounded by the entropy of the discrete probability distribution of $\hat{y}$, $d$ is the reconstruction distortion, $\lambda$ controls the trade-off. The framework is first proposed by Ballé in reference \cite{Bella2016}, where they introduced the widely used generalized divisive normalization (GDN) transform. In their following works, they proposed the hyper prior \cite{Bella2018} to better exploit the probability distribution of the latent feature. And in \cite{BellaCTX} and \cite{LeeCTX}, spacial context structure is proposed to improve the performance of the entropy module. \\
\indent
On the basis of above works, some researchers proposed to use GAN and some perceptual metrics (e.g. LPIPS) to improve the visual perceptual quality. Rippel \cite{RippelGAN} first introduced the effectiveness of GAN in generating perceptual friendly images at an extremely low bitrate. In the following works, Agustsson \cite{AgustssonGAN}, Akutsu \cite{AkutsuGAN}, Dash \cite{DashGAN} also make use of GAN loss to improve the perceptual quality of reconstruction images. According to rate-distortion-perception theory presented in \cite{RDPtradeoff}, Mentzer combines MSE (Mean squared error), LPIPS and GAN loss to generate images of competitive perceptual quality  \cite{HIFIC}. In this paper, we also use a GAN-based architecture, and what distinguishes us from previous works is that we take the image contents into consideration, and adopt different strategies on different target regions. \\
\indent
Another related topic of the work is content-related image compression. Li introduced a content-weighted importance map to guide the allocation of local bit rate \cite{LiCVPR}. Similarly, Cai proposed a CNN based ROI image compression method to allocate more bits to the ROI areas \cite{CaiROI}.  Besides, Zhang proposed an attention-guided dual-layer image compression, which employs a CNN module to predict those pixels on and near a saliency sketch within ROI that are critical to perceptual quality \cite{ZhangCVPR}. However, most of the existing content-related works need to change the architecture and allocate more bits in important fields, thus increasing the network complexity and reducing the compression efficiency. Different from above, we proposed a content-oriented image compression method by adopting different loss strategies on different areas. The advantage is that our scheme can be applied in most of the previous works without changing the architecture. Besides, the image content masks are learned by the network during the training, so no extra information is needed when encoding or decoding an image.

\section{Method}

\subsection{Framework of our method} 

\begin{figure}[h]
    \centering
    \includegraphics[width=0.9\textwidth]{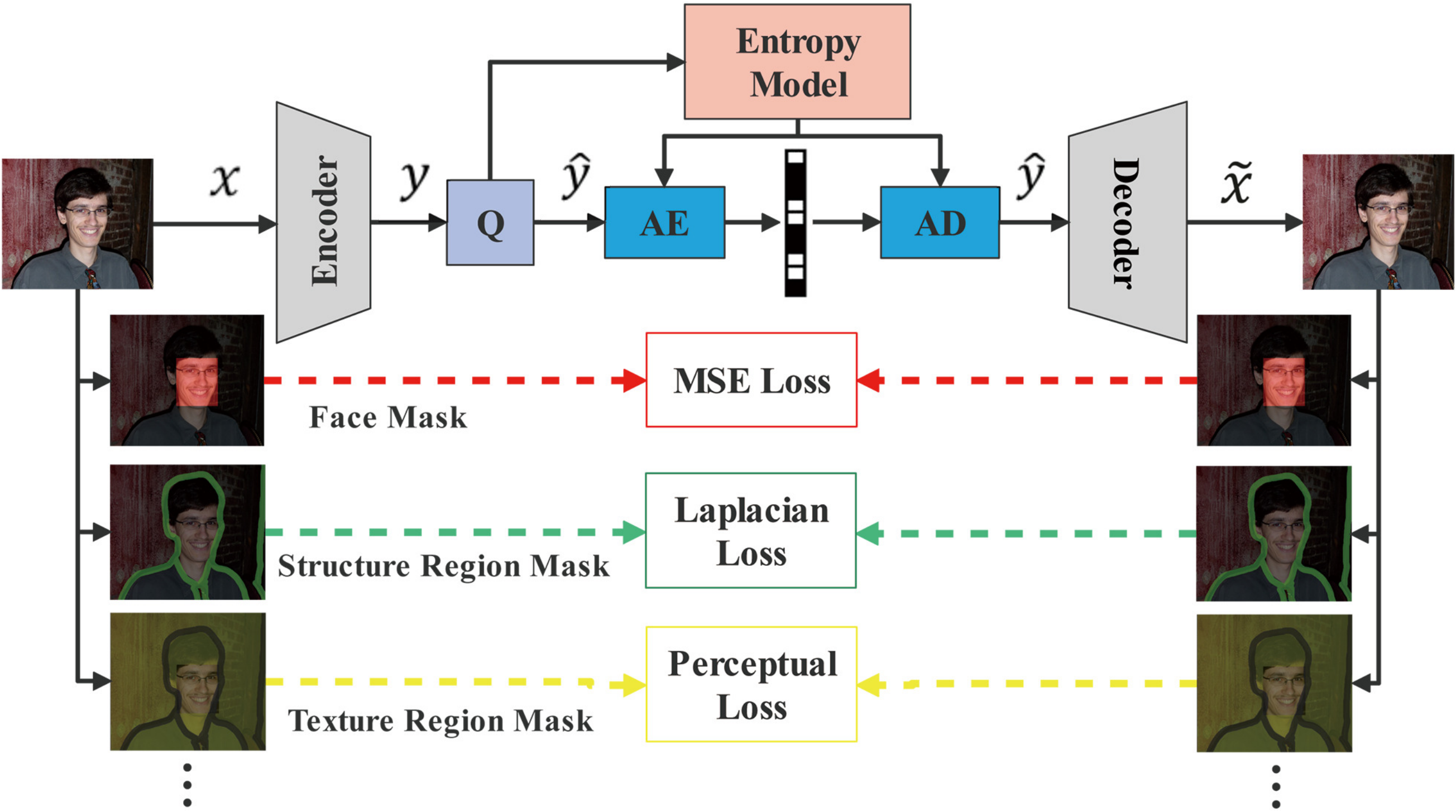}
    \caption{Framework of our method.}
    \label{fig:framework}
\end{figure}

From Section 2.1, it is easy to deduce that different loss functions are suitable for different image contents. However, existing learned image compression always optimizes the whole image with the same loss function. In existing content-related image compression methods, important maps \cite{LiCVPR}, ROI masks \cite{CaiROI}, or other extra information need to be encoded as part of the codestream. One the one hand, such methods require changing the encoder-decoder structure to make the architecture more complex by adding a special module to extract the important maps or ROI masks, on the other hand, they also increase the extra bits that inevitably degrade the coding efficiency. This problem can be overcome if the architecture is ‘clever’ enough to distinguish different image contents and employ different compression strategies. The whole framework of our method is shown in Fig.\ref{fig:framework}. The mask information is only used to select different loss functions during the training phase and not needed in the inference phase.

As shown in Fig.\ref{fig:framework}, E2E image compression is transformed from unlabeled to labeled learning. The choice of the loss functions depends on the classification of image regions. According to the previous analysis, we divide the image into three image regions, namely texture region, structure region, and small face region. 

\subsubsection{Structure Region}
\begin{figure}[h]
    \centering
    \includegraphics[width=0.6\textwidth]{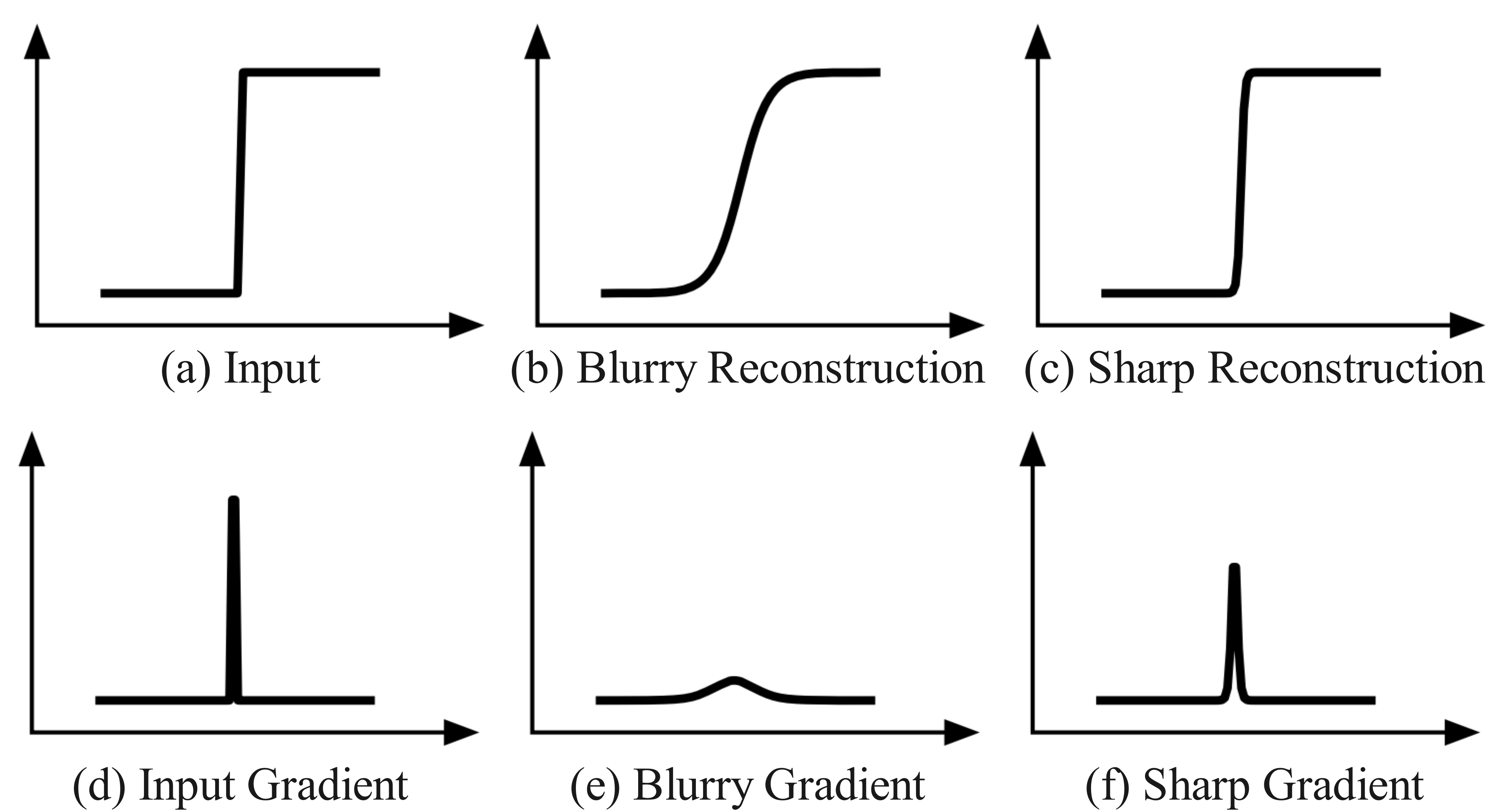}
    \caption{An illumination of a simple 1-D case. The first row shows the pixel sequences and the second row shows their corresponding gradients. }
	\label{grad}
\end{figure}

Structure regions usually have strong gradients and have little statistical correlation with very close neighborhoods due to the abrupt changes in pixels. And loss function with large receptive field will introduce additional noise, which is not acceptable for a precise edge reconstruction. The human eye is sensitive to the sharpness and the pixel-wise correctness of the structure, so a point-wise loss function is more suitable. However, MSE will lead to a blurred reconstruction, affecting the subjective visual perception. As shown in Fig.\ref{grad} \cite{stru}, considering a simple 1-dimensional case. If the model is only optimized in image space by the L1 or MSE loss, we usually get a reconstructed sequence as Fig.\ref{grad}(b) given an input testing sequence whose ground-truth is a sharp edge as Fig.\ref{grad}(a). The model fails to recover sharp edges for the reason that the model tends to give a statistical average. If a second-order gradient constraint is added to the optimization objective as Fig.\ref{grad}(f), the probability of recovering Fig.\ref{grad}(c) is increased significantly. As a results, we choose the Laplacian loss in structure region. The structure loss function is as follows:

\begin{equation}
    \label{ML}
    \mathcal{L}_{stru} = M_{stru} \circ \mathcal{L}_{{lap }}
\end{equation}

\noindent where $M_{stru}$ denotes the mask of the structure region, $\mathcal{L}_{{lap }}$ is computed with Eq.\ref{lap}. We use Laplacian edge detector to generate the $M_{stru}$, which will be introduced in Section 4.1.

\subsubsection{Texture Region}
 Unlike structure regions, texture regions always have more details and domain pixels are highly correlated. It is difficult for the human eyes to perceive pixel-by-pixel correctness, and people are more concerned about the distribution of the texture. Existing perceptual optimization based methods \cite{HIFIC} have achieved very compelling texture reconstruction results. Therefore, we use the perceptual loss in this region. In our work, the perceptual loss are constituted of three part: MAE loss, LPIPS loss, and GAN loss. The formulation can be denoted by:

\begin{equation}
    \label{MP}
    \mathcal{L}_{per} = M_{tex} \circ  \left(\alpha \cdot\mathcal{L}_{{mae}}+\beta \cdot \mathcal{L}_{{LPIPS}}+ \delta \cdot \mathcal{L}_{{GAN}}\right)
\end{equation}

\noindent where $ M_{tex}$ denotes the mask of the texture region, $\mathcal{L}_{{mae}}$, $\mathcal{L}_{{LPIPS}}$ are computed with Eqs.\ref{mse}, \ref{lpips}, $\mathcal{L}_{{GAN}}$ will be introduced in Section 3.2. In our work, the VGG \cite{VGG} network is chosen as the backbone to compute the LPIPS loss.

\subsubsection{Small Face Region}
In our framework, the facial regions will be classified as texture regions and optimised with perceptual loss if not intervened.  In this case, larger faces can be well reconstructed, but small faces may be warped, as shown on the left side of Fig.\ref{fig:intro}.  Therefore, we adopt a different loss function for regions of the small faces. Generally, people are very sensitive to the correctness of the face structure, for which an accurate reconstruction is necessary. Therefore, we use a stricter constraint loss, the MSE loss, for the facial image reconstruction.

\begin{equation}
    \label{MF}
    \mathcal{L}_{sface} = M_{sface} \circ \mathcal{L}_{{mse}}
\end{equation}

where the $M_{sface}$ denotes the mask of the small face regions, and $\mathcal{L}_{{mse}}$ is computed with Eq.\ref{mse}. We use the well-known YOLO-face to detect the faces in the image, and $\mathcal{L}_{sface}$ is only adopted to small faces. 
The bitrate of the quantized latent reprensentation $\hat{y}$ is estimated by the entropy module denoted by P, $R(\hat{y})=-\log (P(\hat{y}))$. Finally, the loss function of the whole image is summaried as:

\begin{equation}
    \label{L}
    \mathcal{L}_{total} = \eta R(\hat{y}) + \epsilon \mathcal{L}_{stru} +  \mathcal{L}_{per} + \gamma \mathcal{L}_{sface}
\end{equation}

where $\eta$, $\epsilon$ and $\gamma$ are weights of corresponding loss metrics. Because people usually pay more attention to faces of the images, we intentionally allocate more bits on small faces by using a larger $\gamma$.
Note that the masks are binary-valued and mutually exclusive. And the priority of the masks is different, specifically, from high to low: the facial mask, the structure mask and the texture mask. In other words, the structure mask doesn’t cover the facial areas, and the texture mask doesn’t cover the facial and structure areas. 

\subsection{Architecture}
\begin{figure}[h]
    \centering
    \includegraphics[width=\textwidth]{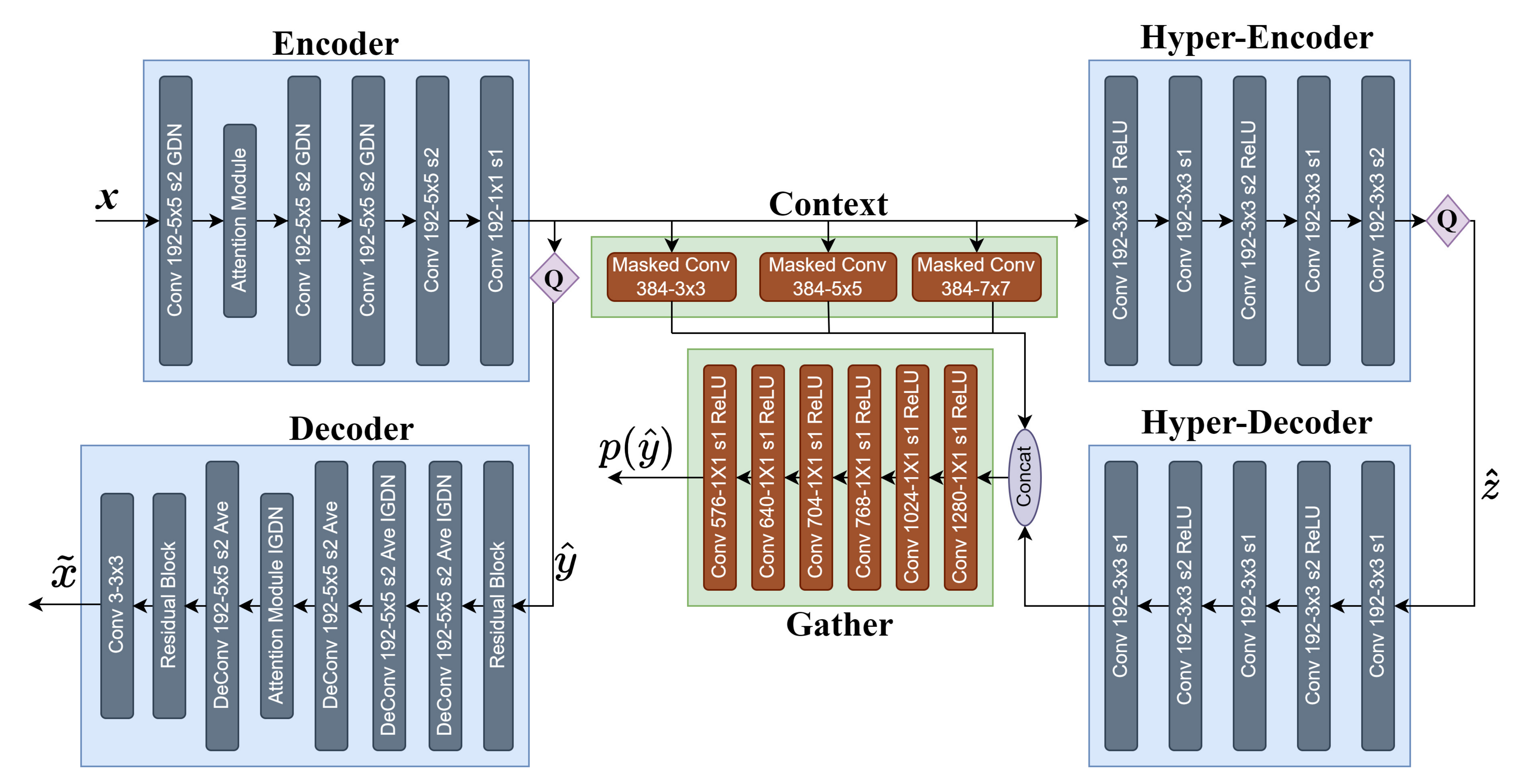}
    \caption{The architecture of our method. $Conv192- 5\times5$ is a convolution with 192 output channels, with $5\times5$ filters. $DeConv$ is a deconvolution operation. $s2$ means the stride of this convolution or deconvolution is 2. $Ave$ means average pooling. $GDN$ or $ReLU$ is used to increase the non-linearity.}
    \label{architecture}
\end{figure}

To prove the effectiveness of our method, we design an E2E image compression architecture, as shown in Fig.\ref{architecture}. It consists of four parts: the encoder E, the decoder/generator G, the entropy module P, and the discriminator D.

\subsubsection{Autoencoder} For the encoder and generator/decoder, GDN \cite{Bella2016} is used to normalize the intermediate feature and also play a role of non-linearity element. Besides, to capture the global dependencies between features, we introduced the attention module (the residual non-local attention block, RNAB) \cite{RNAB} integrated into the architecture. It is well-known that deconvolution operation, or transposed convolution always generates checkerboard artifacts on the reconstructed image. The main reason is, the up-sampled feature map generated by deconvolution can be considered as the result of periodical shuffling of multiple intermediate feature maps computed from the input feature map by independent convolutions \cite{Deconv}. As a result, adjacent pixels on the output feature map are not directly related, resulting in checkerboard artifacts. To alleviate this issue, we add an extra average pooling layer after every deconvolution layer to strengthen the association between adjacent pixels. 
\subsubsection{Entropy model} We adopt the context-based model \cite{BellaCTX} to extract side information $z$, which is used to model the distribution of latent representation $\hat{y}$. And uniform noise U(-1/2,1/2) is used to simulate quantization in the hyper-encoder and when estimating $p(\hat{y}|z)$. The distribution of $\hat{y}$ is modeled with an asymmetric Gaussian entropy model \cite{Cui_2021_CVPR}, which can be denoted by:
\begin{equation}
p_{\hat{y} \mid \hat{z}}(\hat{y} \mid \hat{z}) \sim N\left(\mu, \sigma_{l}^{2}, \sigma_{r}^{2}\right)
\end{equation}
where $\mu$ represent the mean of the latent representation, $\sigma_{l}^{2}$ and $\sigma_{r}^{2}$ represent the estimated left-scale and right-scale parameter, respectively. The asymmetric Gaussian model has stronger representation ability when the entropy estimation do not obey the Gaussian distribution. 

\subsubsection{Discriminator} 
\begin{figure}[h]
    \centering
    \includegraphics[width=0.6\textwidth]{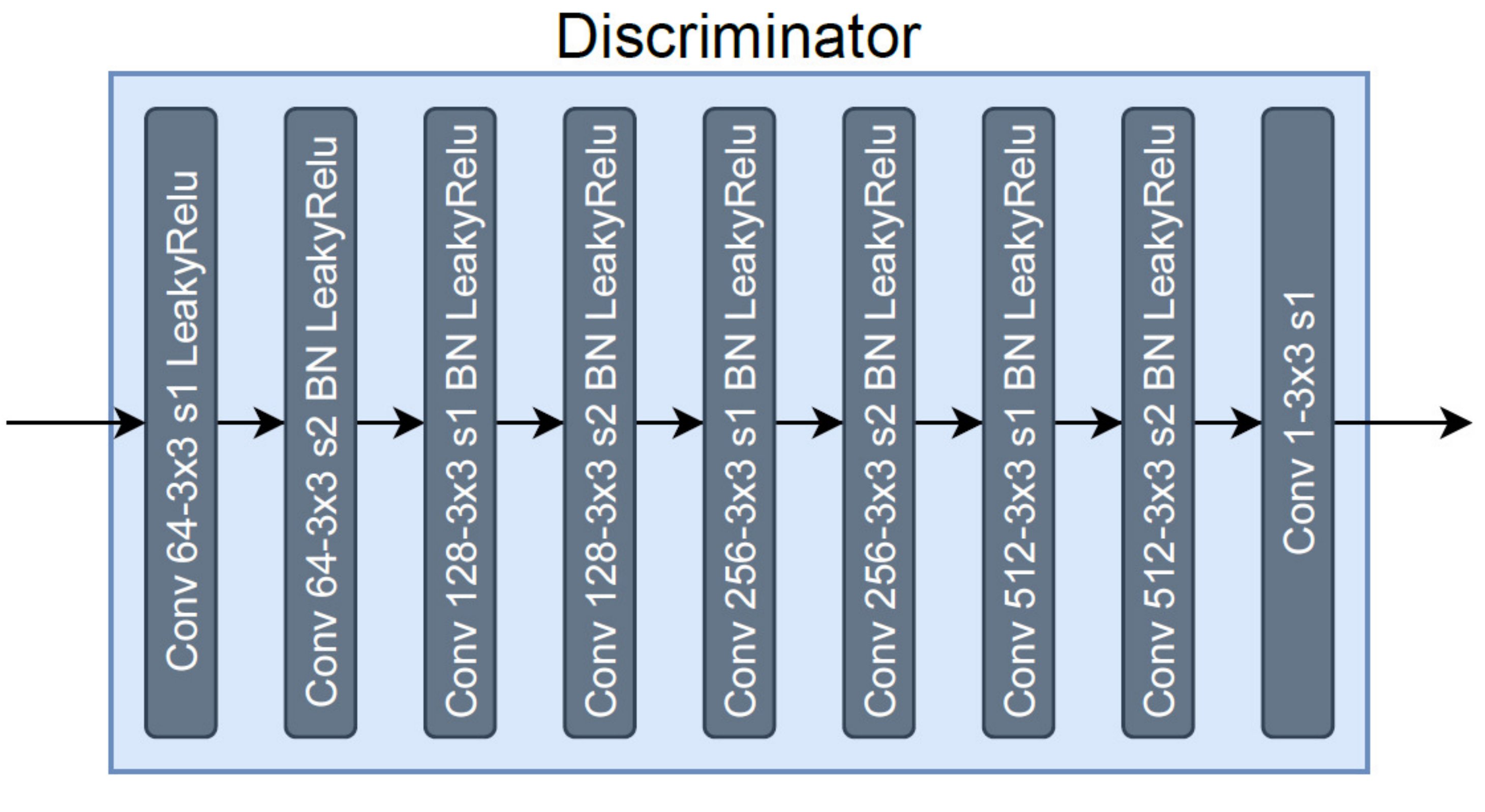}
    \caption{The architecture of discriminator. Same notation is used as in Fig.\ref{architecture}.}
    \label{discriminator}
\end{figure}

In our framework, adversarial training is adopted to improve the perceptual quality of the reconstructed images. Instead of a standard discriminator, we borrow the relativistic average discriminator \cite{RaGAN} used in \cite{ESRGAN}, which tries to predict the probability that a groundtruth image $x$ is relatively more realistic than a generated one $\tilde{x}$, on average. The loss is divided into two parts, the generator loss $\mathcal{L}_{D}^{R a}$, and the discriminator loss $\mathcal{L}_{G}^{R a}$ :
\begin{equation}
\label{Ragan}
\begin{aligned}
    \mathcal{L}_{D}^{R a}=-\mathbb{E}_{x}\left[\log \left(D_{R a}\left(x, \tilde{x}\right)\right)\right]-\mathbb{E}_{\tilde{x}}\left[\log \left(1-D_{R a}\left(\tilde{x}, x\right)\right)\right]\\
    \mathcal{L}_{G}^{R a}=-\mathbb{E}_{x}\left[\log \left(1-D_{R a}\left( x,\tilde{x}\right)) \right]-\mathbb{E}_{\tilde{x}}\left[\log \left(D_{R a}\left(\tilde{x},x \right))\right]\right.\right.
\end{aligned}
\end{equation}
where $D_{R a}\left(x, \tilde{x}\right)=\sigma\left(C\left(x\right)-\mathbb{E}_{\tilde{x}}\left[C\left(\tilde{x}\right)\right]\right)$. $C\left(x\right)$ and $C\left(\tilde{x}\right)$ are the the non-transformed discriminator ouput. $\sigma$ is the sigmoid function, and $\mathbb{E}_{x}[\cdot]$ computes the average output. Moreover, the PatchGAN discriminator \cite{PatchGAN}, which has been proven to improve the quality of the generated images, is also utilized in our architecture. The PatchGAN has fewer parameters. And it not only preserves more texture but can also be applied to images of arbitrary sizes. The detailed architecture of our discriminator is shown in Fig.\ref{discriminator}.

\subsection{Implementation of masked perceptual loss}
Usuallym, Adding a pixel-level mask to MSE or MAE is easy with simple point-wise multiplication. But it is a little harder to used on LPIPS or GAN losses, because these two loss functions compute the feature losses and cannot correspond to mask pixel-to-pixel. In this section, we propose a method called ‘real value replacement’ to solve the problem.\\
\indent Consider the mask $M_{tex}$, the original image $x$, and the reconstructed image $\tilde{x}$, we replace the value of the mask part of $\tilde{x}$ with the corresponding value of $x$ to get the replaced reconstructed image $\tilde{x}'$:
\begin{equation}
    \tilde{x}' = \left( 1 - M_{tex}\right) \circ x + M_{tex} \circ \tilde{x}
\end{equation}
Then calculate the loss function directly with $x$ and $\tilde{x}'$ to estimate the masked loss:
\begin{equation}
\begin{aligned}
    M_{tex} \circ \mathcal{L}_{{LPIPS}} \left(x, \tilde{x}\right) &\approx \mathcal{L}_{{LPIPS}} \left(x, \tilde{x}'\right) \\
	 M_{tex} \circ \mathcal{L}_{{GAN}} \left(x, \tilde{x}\right) &\approx \mathcal{L}_{{GAN}} \left(x, \tilde{x}'\right)
\end{aligned}
\end{equation}

\section{Experiment}
\subsection{Training Details}
Due to the adoption of face loss, a large number of face images are required for training. For this purpose, we use the well-known MSCOCO dataset \cite{MSCOCO} as our training set, and Kodak dataset \cite{Kodak} as our testing set. It is notoriously hard to train GANs for image generation, so the training procedure is divided into two stages. In the first stage, we only use the MSE as the distortion loss to guide the optimization at the pixel level reconstruction. The optimization target in the first stage is to minimize $\mathcal{L}_{R D}=\eta R+ \mathcal{L}_{mse}$. Using the first stage result as the pre-trained model, we can train the perceptual optimized model with the loss function Eq.\ref{L} mentioned in section 3.2 in the second stage. During the training phase, we randomly crop the images into patches of size $256\times256$ and set the batch size to 8. The initial learning rate is set to 0.0001 and halved at 160k and 500k iteration. We use kaiming initialization \cite{INIT} to initialize all our models. The weights of the loss are set: $\alpha = 0.01$, $\beta = 1$, $\gamma = 0.2$, $\delta = 0.0005$, and $\epsilon=0.3$. We only modify $\eta$ for different target bit-rates.\\
\indent
In order to use different loss metrics on different regions of the images, the masks of different content regions are required. In our experiment, the faces in the images are detected using the well-known YOLO-face \cite{YOLO}. Then the coordinate information of the faces is stored in an XML file, which is used to generate the face masks in the training phase. By doing this, we can save the face detection time during training the model. For structure region masks, due to the low complexity of edge detection, we do not generate masks in advance, but directly use Laplacian edge detectors to detect structure regions during training.

\begin{figure}[h]
    \centering
    \centering
    \includegraphics[width=0.9\textwidth]{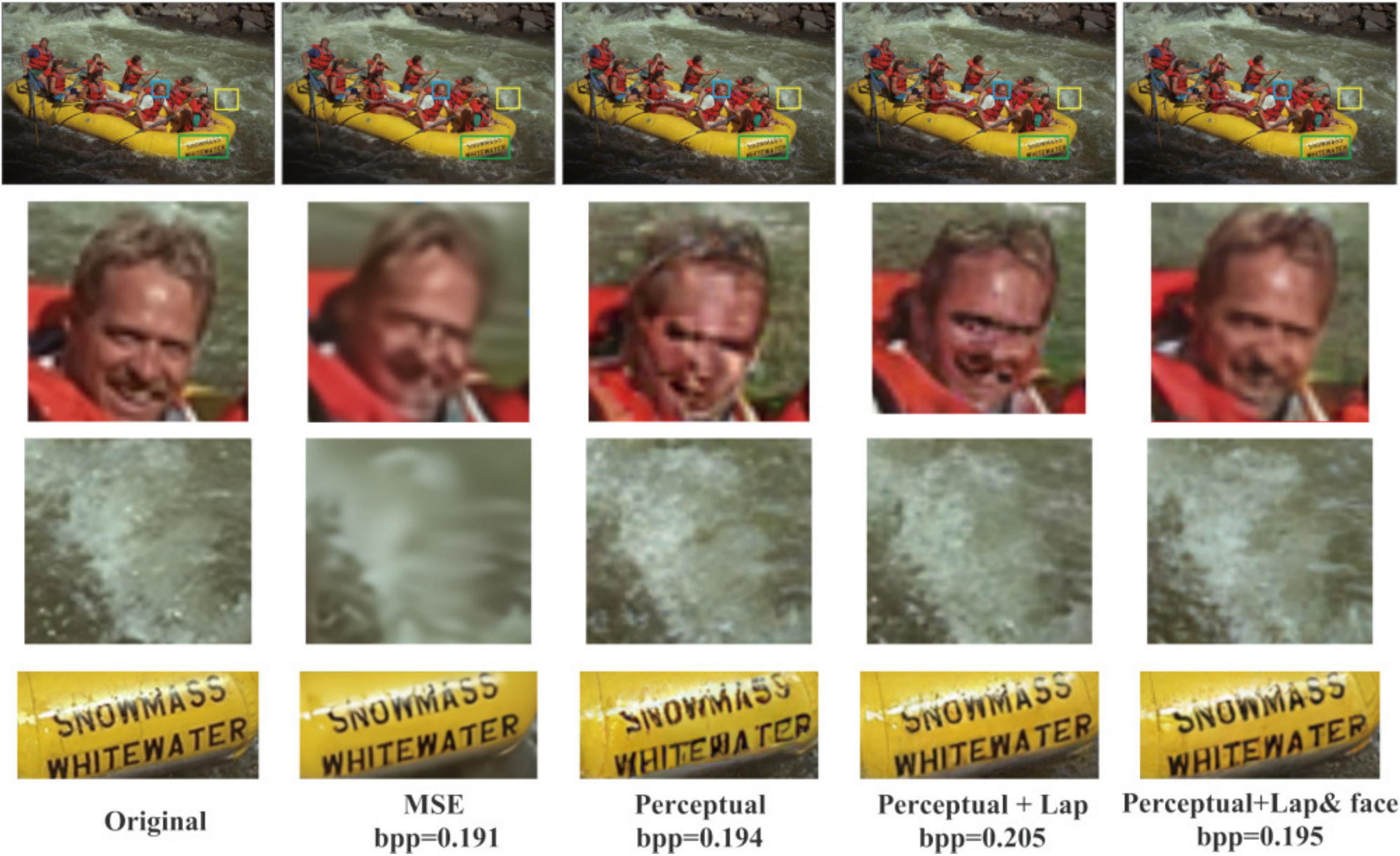}
    \caption{Comparison of different loss strategies.}
    \label{fig:ablation}
\end{figure} 

\begin{figure}[h]
\begin{minipage}[t]{0.5\linewidth}
\centering
\includegraphics[width=0.9\linewidth]{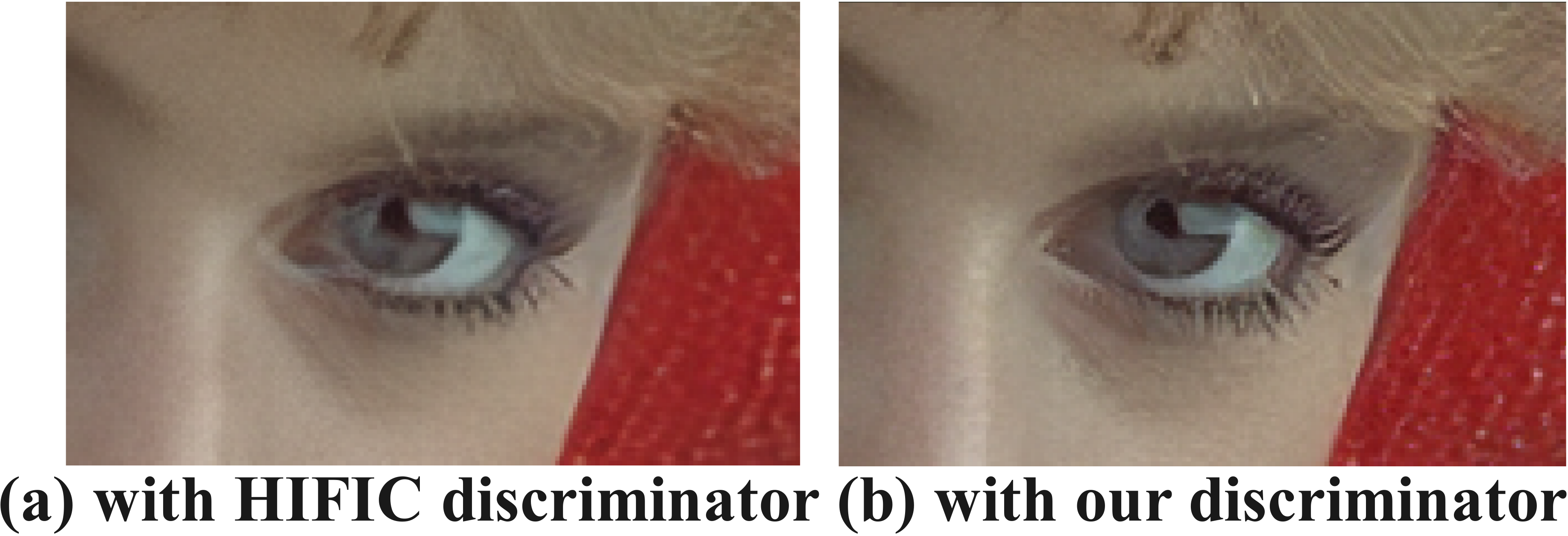}
\caption{Comparison of the discriminators.}
\label{fig:cmp_disc}
\end{minipage}%
\begin{minipage}[t]{0.5\linewidth}
\centering
\includegraphics[width=0.9\linewidth]{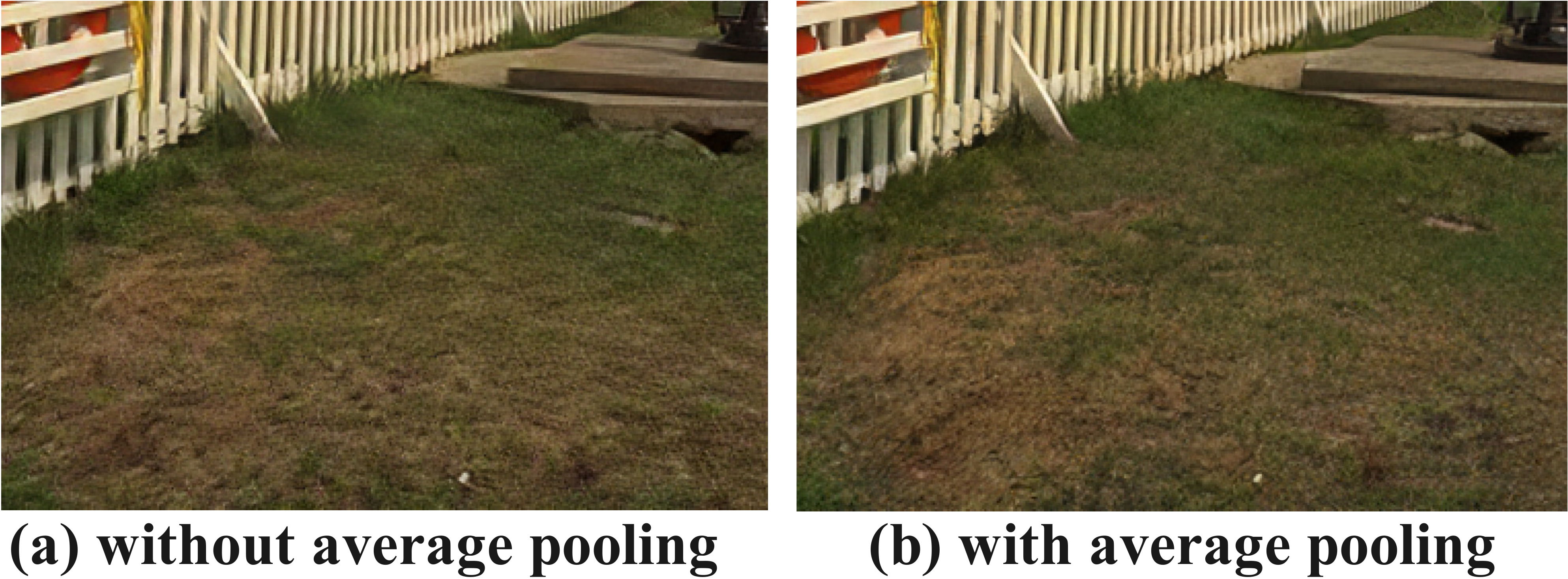}
\caption{Ablation on the average pooling.}
\label{fig:ave}
\end{minipage}
\end{figure}
\subsection{Ablation Study}
\subsubsection{Ablation on the loss metric}
In order to show the effectiveness of the proposed, we conduct several comparative experiments as follows.  
\begin{itemize}
\setlength{\itemsep}{0pt}
\setlength{\parsep}{0pt}
\setlength{\parskip}{0pt}
\item Case 1, the model is optimized only with MSE loss. (MSE) 
\item Case 2, the model is optimized with MAE, LPIPS and GAN loss. (Perceptual)
\item Case 3, the model is optimized with MAE, LPIPS, GAN and Laplacian loss. (Perceptual + Lap)
\item Case 4, the model is optimized with MAE, LPIPS, GAN, Laplacian and face loss. (Perceptual + Lap \& face)
\end{itemize}
Take the image kodim14 an example, the visual comparison of different cases is shown in Fig.\ref{fig:ablation}. All of these cases are compressed at a similar bitrate, around 0.2 bpp. Overall, it can be clearly observed that the reconstructed image in case 5 has high-fidelity facial features, clear and correct text, informatively detailed backgrounds and an overall harmonious visual quality. Specifically, in case 1, the faces and water waves are too blurred, which attributes to the property of the MSE loss function. In case 2, with the help of additional loss functions GAN and LPIPS, the informative details of water waves are restored, but the facial features and text are severely warped. In case 3, with the help of additional edge loss, not only the informative details of water waves are restored, but the text distortion phenomenon also disappears. However, the facial distortion remains. Until all the losses are added, in case 5, the best visual effect is achieve in all the areas. Through the ablation study, we can draw two conclusions: (1) the architecture proposed in 3.3 is able to distinguish different image contents; (2) adopting appropriate loss function for different image content helps improve the visual quality of reconstructed images.

\begin{figure}[h]
    \centering
    \includegraphics[width=0.9\textwidth]{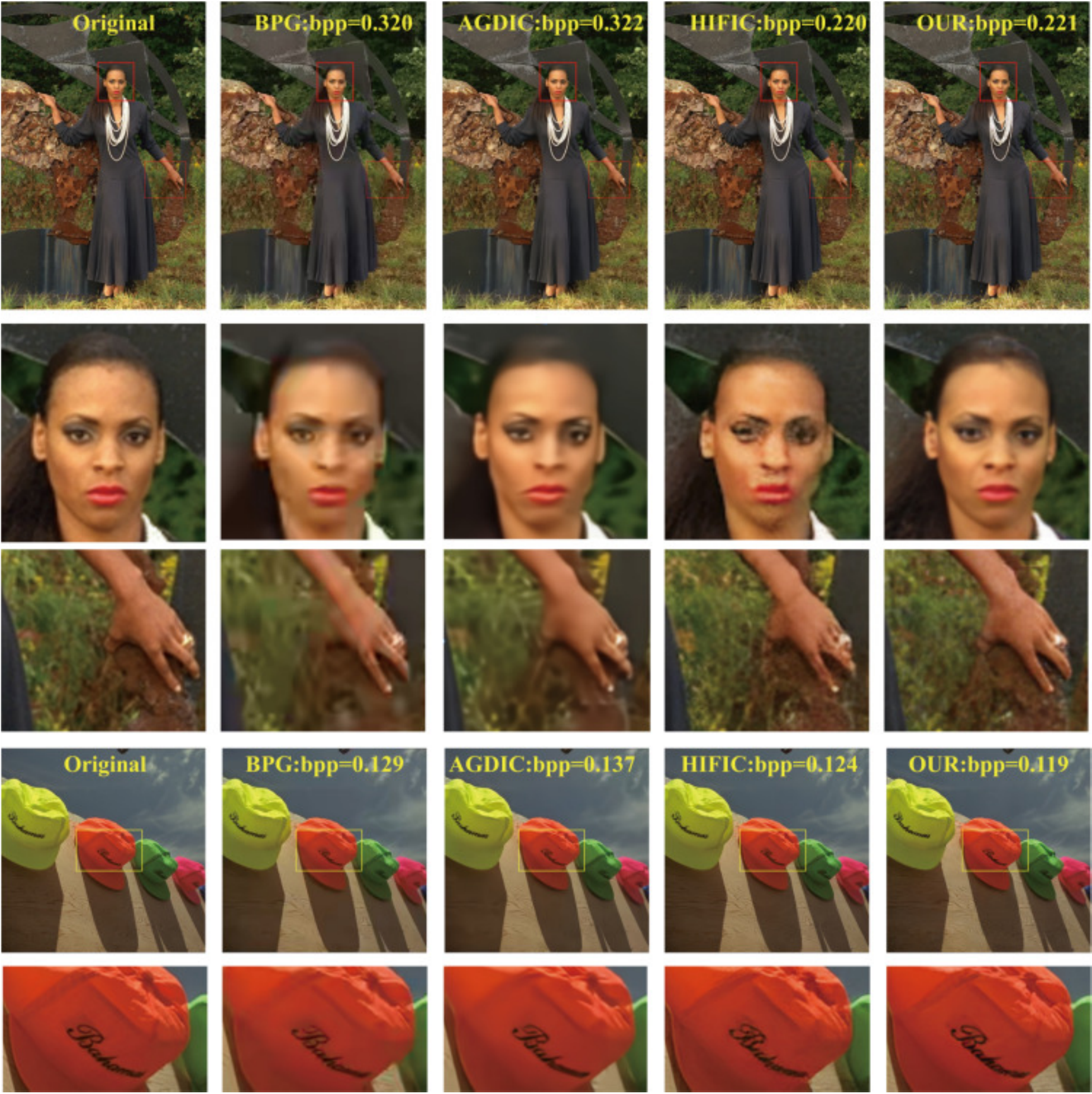}
    \caption{Compressed results of kodim18 using different methods.}
    \label{fig:comparison}
\end{figure}

\subsubsection{Ablation on the architecture}
In order to get better result, we choose the better components in our architecture, such as the asymmetric Gaussian entropy model, the relatively average discriminator,and the average pooling. We have done relevant experiments to demonstrate the effectiveness of our architecture. The results show that with the help of the asymmetric Gaussian model, we achieve about 0.68\% BD-rate reduction compared with the general Gaussian entropy model. We also tested the effectiveness of our disriminator, and the result shown in Fig.\ref{fig:cmp_disc} proves that our discriminator can preserve more details than that of HIFIC. Besides, Fig.\ref{fig:ave} shows that the average polling can alleviate the checkerboard artifacts.

\subsection{Visual Results}
\begin{figure}[H]

    \centering
    \centering
    \includegraphics[width=0.9\textwidth]{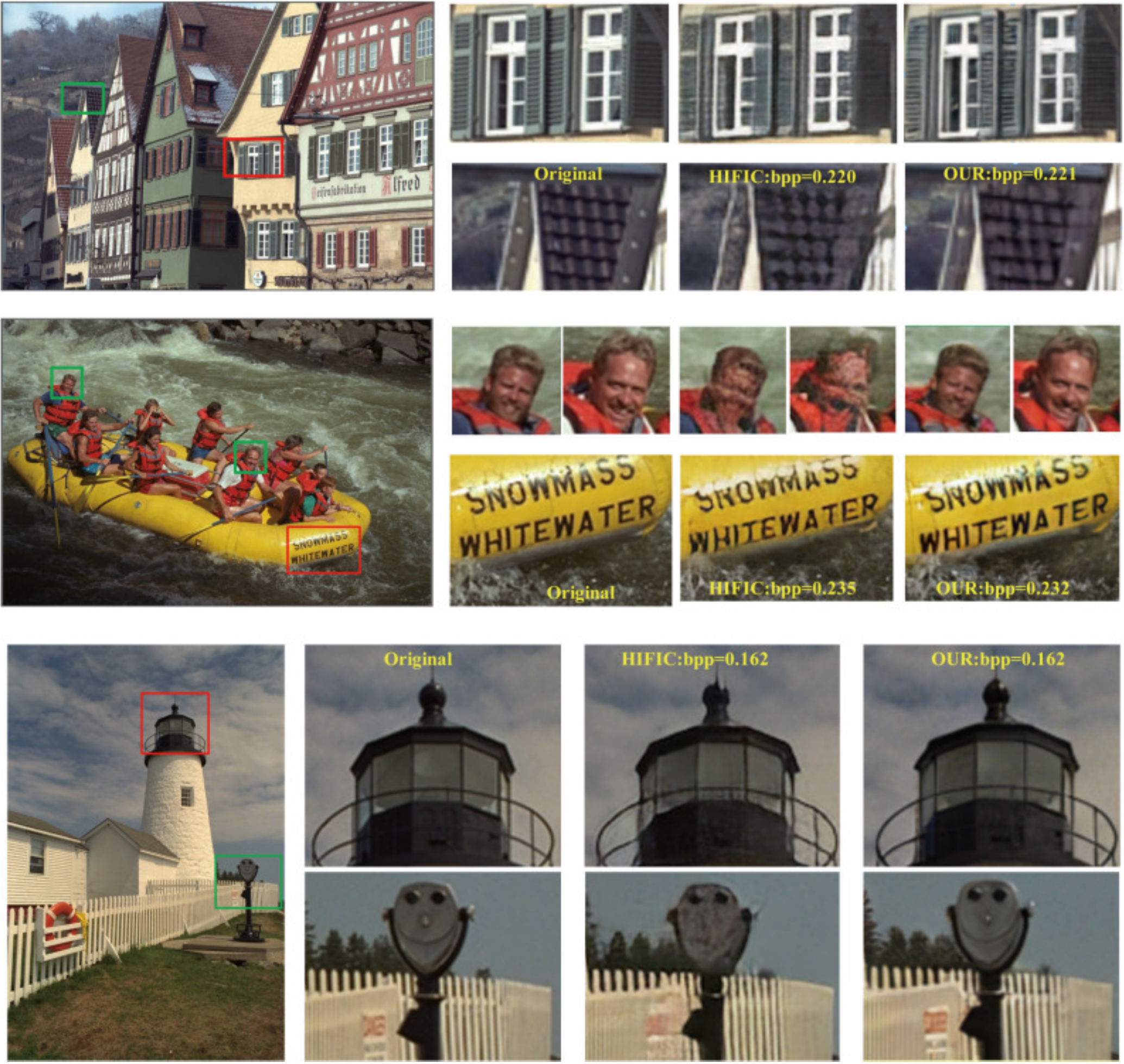}
    \caption{Compressed results of kodim18 using different methods.}
    \label{fig:comparison_hific}
\end{figure}

To show the advantage of our method, we compare our method (named COLIC) with BPG, Asymmetric Gained Deep Image Compression (AGDIC) optimized with MSE \cite{Cui_2021_CVPR} and High-Fidelity Generative Image Compression (HIFIC, SOTA perceptual optimized method) \cite{HIFIC}.Since BPG and AGDIC are not perceptual-optimized, we select relatively higher bitrates for them. The visual results are shown in Fig.\ref{fig:comparison}, in which the cropped blocks highlight the reconstruction of certain regions, such as textures, texts, and small faces. As can be seen from Fig.\ref{fig:comparison}, although BPG and AGDIC can maintain the correct facial structure and achieve relatively better results in text area, the texture results are over-smoothed. As shown in the third row, the background generated by BPG and AGDIP is too blurry. On the contrary, although HIFIC can reconstruct the informative details, it will lead to the distortion of face and text. As shown in the second and fifth rows, the face and text generated by HIFIC are overwarpped. Compared to these results, it is clear that our proposed COLIC achieves the best visual effects. It handles all these situations better, recovering more texture detail, correct small face and text structure.\\
\indent More examples of comparison with HIFIC are shown in Fig.\ref{fig:comparison_hific}. Compared with HIFIC, COLIC can not only recover similar informative texture regions, such as the spray, but also better structured regions such as lines, textures and small faces. In a word, COLIC can achieve better visual effects. 

\begin{figure}[h]
    \centering
    \centering
    \includegraphics[width=\textwidth]{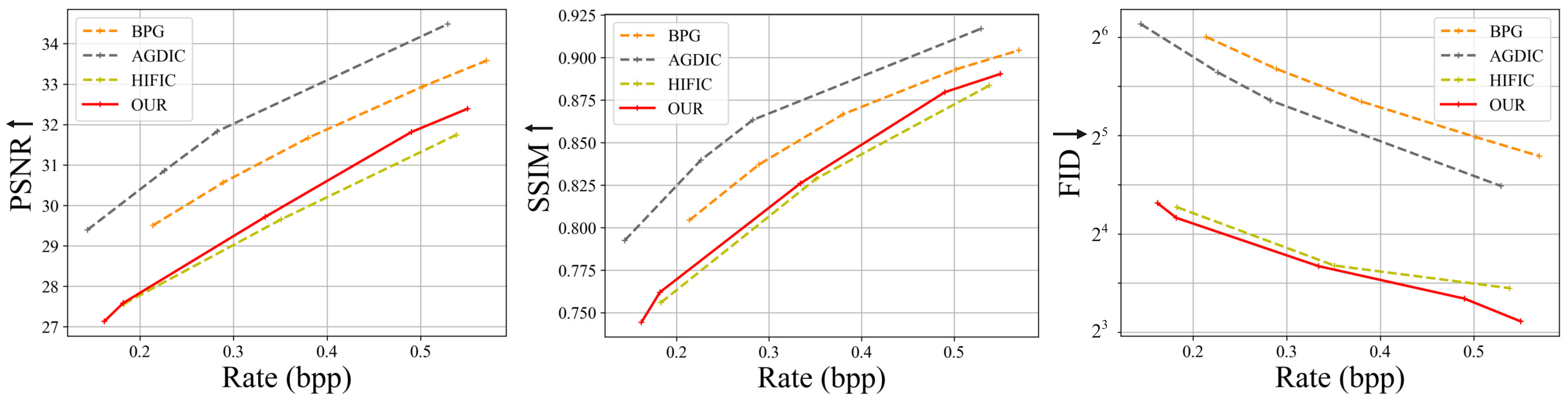}
    \caption{Rate-distortion and -perception curves on Kodak. Arrows indicate whether lower is better ($\downarrow$), or higher is better ($\uparrow$).}
    \label{fig:metrics}
\end{figure}

\subsection{Statistics result}
In this section, we use metrics such as PSNR, SSIM, and FID to quantify our results. PSNR and SSIM are widely used in various digital image evaluations, which compute the average pixel-wise distance and structural similarity between the two images. Unlike PSNR, SSIM, which measure the similarity between individual images pairs, FID assesses the similarity between the distribution of the reference images and the generated/distorted images. FID is a good metric for evaluating the subjective results and is widely used for super-resolution or generative tasks.  \\
\indent
The R-D curves of COLIC, BPG, AGDIC and HIFIC are shown in Fig.\ref{fig:metrics}.  Compared with BPG and AGDIC, as expected, COLIC and HIFIC dominate in perceptual metric FID, but relatively poor on objective metrics PSNR and SSIM, which can be explained by the rate-distortion-perception theory \cite{RDPtradeoff} that they sacrifice some objective performance to improve the perceptual quality. In order to improve the visual quality, we sacrifice some of the objective metrics. Compared with HIFIC, COLIC achieves better results on both objective and perceptual metrics. It attributes to the better optimization loss that imposes the correct constraints on the structure region, the better network structure that with better reconstruction ability, and better training strategy.

\section{Conclusion}
In this work, we propose a content-oriented image compression scheme, which can be used in most existing methods. We suggest that different loss metrics should be used on different image contents according to their characteristics. And a GAN-based architecture is designed to prove the effectiveness of our scheme. Experiments clearly show the superiority of our method on visual quality as well as different metrics. In fact, the effectiveness of this method demostrates that existing encoders and decoders are 'smart' enough to distinguish different image regions and employ different reconstruction strategies with the guidance of different training loss. Therefore, human perceptual priors can be better utilized through supervised training methods to obtain better subjective results.

\clearpage
%
%
\bibliographystyle{splncs04}
\bibliography{reference}
\end{document}